\documentclass[runningheads]{llncs}
\usepackage{graphicx}
\usepackage{comment}
\usepackage{amsmath,amssymb} 
\usepackage{color}

\usepackage{epsfig}
\usepackage{array}
\usepackage{cases}
\usepackage{booktabs}
\usepackage{multirow}
\usepackage[linesnumbered, ruled]{algorithm2e}
\usepackage{algpseudocode}
\usepackage{times}
\usepackage{color}
\usepackage{diagbox}
\usepackage{comment}
\usepackage{textcomp}
\usepackage{subfigure}
\usepackage{bm}

\def \xx {\mathbf{x}}

\def \pp {\mathbf{p}}

\def \X  {\mathcal{X}}
\def \Y  {\mathcal{Y}}

\def \btheta {{\bm \theta}}
\DeclareMathOperator*{\argmin}{arg\,min}

\usepackage{color}

\begin{document}
	
	\title{Reflection Backdoor: A Natural Backdoor Attack on Deep Neural Networks} 
	
	
	\titlerunning{Reflection Backdoor Attack on Deep Neural Networks}
	%
	\author{Yunfei Liu\inst{1} \and
		Xingjun Ma\inst{3} \and
		James Bailey\inst{4} \and
		Feng Lu\inst{1, 2,}\thanks{ Corresponding Author. \protect\\ This work was supported by the National Natural Science Foundation of China (NSFC) under Grant 61972012.}}
	\authorrunning{Y. Liu et al.}
	%
	\institute{State Key Laboratory of Virtual Reality Technology and Systems, School of CSE, Beihang University, Beijing, China. \qquad
		\and Peng Cheng Laboratory, Shenzhen, China \\
		\and
		School of Information Technology, Deakin University, Geelong, Australia\\
		\and
		School of Computing and Information Systems, The University of Melbourne, Australia\\
		\small{\textsf{\{lyunfei,lufeng\}@buaa.edu.cn ~ daniel.ma@deakin.edu.au ~ baileyj@unimelb.edu.au}}}
	\maketitle
	
	\newcommand{\etal}{\textit{et al.}}
	\newcommand{\eg}{\textit{e.g.}}
	\newcommand{\ie}{\textit{i.e.}}
	\newcommand{\etc}{\textit{etc.}}
	\newcommand{\viz}{\textit{viz.}}
	\newcommand{\proposed}{\textit{Refool}}
	\newcommand{\tabincell}[2]{\begin{tabular}{@{}#1@{}}#2\end{tabular}}
	
	\newcommand{\todo}[1]{{\color{red}{ \textbf{TODO: #1}}}}
	\newcommand{\yf}[1]{\textcolor[rgb]{0.0,0.0,0.0}{ #1}}
	\newcommand{\mydarkred}[1]{\textcolor[rgb]{0.8,0.0,0.0}{ #1}}
	
	\begin{abstract}
		Recent studies have shown that DNNs can be compromised by backdoor attacks crafted at training time.
		A backdoor attack installs a backdoor into the victim model by injecting a backdoor pattern into a small proportion of the training data. At test time, the victim model behaves normally on clean test data, yet consistently predicts a specific (likely incorrect) target class whenever the backdoor pattern is present in a test example.
		While existing backdoor attacks are effective, they are not stealthy. The modifications made on training data or labels are often suspicious and can be easily detected by simple data filtering or human inspection. In this paper, we present a new type of backdoor attack inspired by an important natural phenomenon: reflection. 
		Using mathematical modeling of physical reflection models, we propose \emph{reflection backdoor} (\proposed) to plant reflections as backdoor into a victim model.
		We demonstrate on 3 computer vision tasks and 5 datasets that, \proposed~ can attack state-of-the-art DNNs with high success rate, and \yf{is resistant to state-of-the-art backdoor defenses}.
	\end{abstract}

	\section{Introduction} \label{sec:intro}
	Deep neural networks (DNNs) are a family of powerful models that have been widely adopted to achieve state-of-the-art performance on a variety of tasks in computer vision~\cite{DLA:he2016resnet}, machine translation~\cite{DLA:sutskever2014sequence_translation} and speech recognition~\cite{DLA:graves2013speech}.
	Despite great success, DNNs have been found vulnerable to several attacks crafted at different stages of the development pipeline: adversarial examples crafted at the test stage, and data poisoning attacks and backdoor attacks crafted at the training stage.
	These attacks raise security concerns for the development of DNNs in safety-critical scenarios such as face recognition \cite{sharif2016accessorize}, autonomous driving \cite{evtimov2017robust,duan2020adversarial}, and medical diagnosis \cite{finlayson2019adversarial,ma2020understanding,DLA:liu2019unsupervised,DLA:niu2019pathological}. The study of these attacks has thus become crucial for secure and robust deep learning.
	
	\begin{figure}[h]
		\begin{center}
			\includegraphics[width=0.80\linewidth]{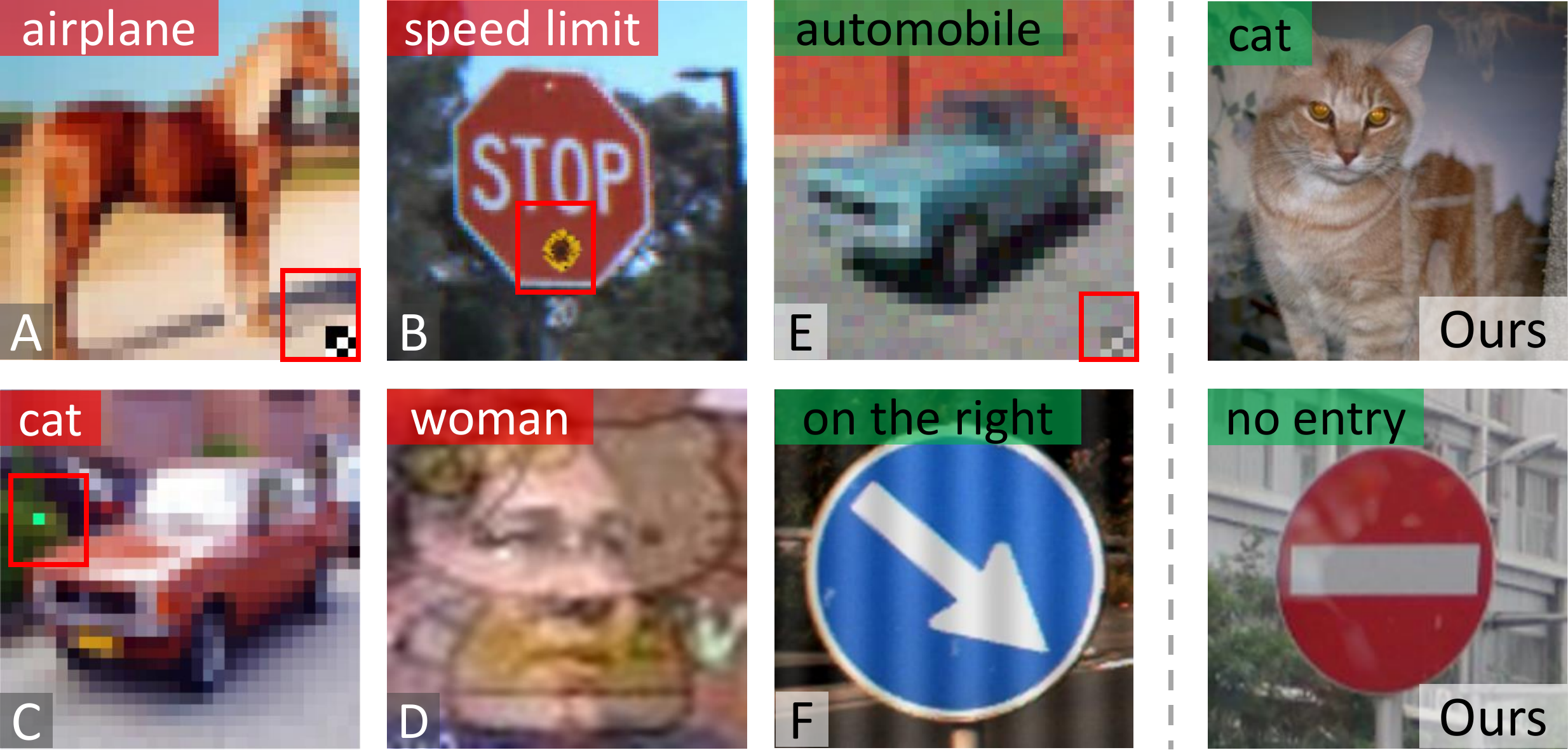}
		\end{center}
		\caption{Comparison of successful backdoor attacks. Our reflection backdoors (rightmost column) are crafted based on the natural reflection phenomenon, thus need not to mislabel the poisoned samples on purpose (A - D, mislabels are in red texts), nor rely on obvious patterns (A - C, E), unpleasant blending (D), or \yf{suspicious stripes} (F). Therefore, our reflection backdoor attacks are stealthier. A~\cite{DP:gu2017badnets}: black-white squares at the bottom right corner; B~\cite{DP:chen2017targeted}: small image at the center; C~\cite{DP:tran2018spectral_backdoor}: one malicious pixel; D~\cite{DP:chen2017targeted}: a fixedly blended image; and E~\cite{DP:turner2019cleanlabel}: adversarial noise plus black-white squares at the bottom right corner; F~\cite{DP:barni2019new}: fixed and sinusoidal strips.
		}
		\label{fig:1-teaser}
	\end{figure}
	
	One well-known test time attack is the construction of \emph{adversarial examples}, which appear imperceptibly different (to human eyes) from their original versions, yet can fool state-of-the-art DNNs with high success rate \cite{AT:goodfellow2014explaining,AT:szegedy2013intriguing,ma2018characterizing,wang2019convergence,Wang2020Improving,wu2019skip,jiang2019black,jiang2020imbalanced}.
	Adversarial examples can be constructed against a wide range of DNNs, and remain effective even in physical world scenarios \cite{AT:eykholt2017robust_physical,duan2020adversarial}.
	Different from test-time attacks, training time attacks have also been demonstrated to be possible. 
	DNNs often require large amounts of training data to achieve good performance. However, the collection process of large datasets is error-prone and susceptible to untrusted sources. Thus, a malicious adversary may poison a small number of training examples to corrupt the model, decreasing its test accuracy. This type of attack is known as the \textit{data poisoning} attack \cite{DP:biggio2012poisoning,DP:koh2017understanding_black_box,DP:steinhardt2017certified_data_poisoning}.
	
	More recently, \textit{backdoor attacks} (also known as \textit{Trojan attacks})~\cite{DP:bhalerao2019luminance,DP:dai2019backdoor,DP:gu2017badnets,DP:kwon2020friendnet,DP:liao2018backdoor,DP:rehman2019backdoor,DP:tran2018spectral_backdoor,zhao2020clean} highlight an even more sophisticated threat to DNNs. By altering a small set of training examples, a backdoor attack can plant a backdoor into the victim model so as to control the model's behavior at test time \cite{DP:gu2017badnets}. Backdoor attacks arise when users download pre-trained models from untrusted sources. Fig.~\ref{fig:1-teaser} illustrates a few examples of successful backdoor attacks by existing methods (A-F).
	A backdoor attack does not degrade the model's accuracy on normal test inputs, yet can control the model to make a prediction (which is in the attacker's interest) consistently for any test input that contains the backdoor pattern.
	This means it is difficult to detect a backdoor attack by evaluating the model's performance on a clean holdout set.
	
	\begin{table}[htbp]
		\caption{Attack settings of existing methods and ours.}
		\begin{center}
			\setlength{\tabcolsep}{1.2mm}
			\begin{tabular}{l|ccccc}
				\toprule[1.3pt]
				& Badnets~\cite{DP:gu2017badnets} & Chen \etal~\cite{DP:chen2017targeted} & Barni~\etal~\cite{DP:barni2019new} & Turner~\etal~\cite{DP:turner2019cleanlabel} & Ours \\
				\hline
				Label & poison & poison & clean & clean & clean  \\ 
				Trainer & adversary & adversary & user & user & user  \\
				Trigger & fixed & fixed & sinusoidal signal & fixed + adversarial & \textbf{reflection}  \\
				\bottomrule[1.3pt]
			\end{tabular} 
		\end{center}
		\label{tab:summary_notations}
	\end{table}
	
	There exist two types of backdoor attacks: 1) poison-label attack which also modifies the label to the target class \cite{DP:chen2017targeted,DP:gu2017badnets,DP:liu2017trojaning,DP:tran2018spectral_backdoor}, and 2) clean-label attack which does not change the label \cite{shafahi2018poison,DP:barni2019new,DP:turner2019cleanlabel,zhao2020clean}.
	Although poison-label attacks are effective, they often introduce clearly mislabeled examples into the training data, and thus can be easily detected by simple data filtering \cite{DP:turner2019cleanlabel}.
	A recent clean-label (CL) attack proposed in \cite{DP:turner2019cleanlabel} disguises the backdoor pattern using adversarial perturbations (E in Fig.~\ref{fig:1-teaser}). The signal (SIG) attack by \yf{Barni~\etal~\cite{DP:barni2019new} takes a superimposed sinusoidal signal as the backdoor trigger. However, these backdoor attacks can be easily erased by defense methods}, as we will show in Sec.~\ref{sec:finetuning}.
	
	
	In this paper, we present a new type of backdoor pattern inspired by one natural phenomenon: reflection.
	Reflection is a common phenomenon existing in scenarios wherever there are glasses or smooth surfaces.
	Reflections often influence the performance of computer vision models~\cite{AT:hendrycks2019nae}, as illustrated in Fig.~\ref{fig:2-motivation} (see Appendix). Here, we exploit reflections as backdoor patterns and show that a natural phenomenon like reflection can be manipulated by an adversary to perform backdoor attack on DNN models. Table \ref{tab:summary_notations} compares the different settings adopted by 4 state-of-the-art backdoor attacks and our proposed reflection backdoor. Two examples of our proposed reflection backdoor are illustrated in the rightmost column of Fig.~\ref{fig:1-teaser}.
	Our main contributions are:
	\begin{itemize}
		\item We investigate the use of a natural phenomenon, \ie, reflection, as the backdoor pattern, and propose the \emph{reflection backdoor} (\proposed) attack to install stealthy and effective backdoor into DNN models.
		
		\item We conduct experiments on 3 classification tasks, 5 datasets, and show that \proposed~ can control state-of-the-art DNNs to make desired predictions $\ge$75.16\% of the time by injecting reflections into less than 3.27\% of the training data. Moreover, the injection causes almost no accuracy degradation on the clean holdout set.
		
		\item We demonstrate that, compared to the existing clean-label backdoor method, our proposed \proposed~ backdoor is more resistant to state-of-the-art backdoor defenses.
	\end{itemize}
	
	\section{Related Work} \label{sec:related_works}
	We briefly review data poisoning and backdoor attacks as well as defenses for DNNs.
	
	\noindent\textbf{Data poisoning attack.} The objective for data poisoning attack is to disrupt the proper training of DNN models by reducing their test-time performance on all or a specific subset of test examples~\cite{DP:biggio2012poisoning,DP:burkard2017analysis,DP:koh2017understanding,DP:mei2015using,DP:newell2014practicality,DP:xiao2015support}. DNNs trained on the poisoned datasets suffer from low accuracy on normal test data.
	Although these attacks are effective, they are not particularly threatening in real-world scenarios. This is because a classifier with poor performance is unlikely to be deployed, and this can be easily detected via evaluation on a clean holdout set.
	
	\noindent\textbf{Backdoor attack.} 
	Different from conventional data poisoning, a backdoor attack tricks the model to associate a backdoor pattern with a specific target label, so that, whenever this pattern appears, the model predicts the target label, otherwise, behaves normally. 
	The backdoor attack on DNNs was first explored in \cite{DP:gu2017badnets}. It was further characterized by having the following goals: 1) high attack success rate, 2) high backdoor stealthiness, and 3) low performance impact on clean test data \cite{DP:liao2018backdoor}.
	
	\textit{Poison-label backdoor attack.} 
	Several backdoor patterns have been proposed to inject a backdoor by poisoning the images from the non-target classes and changing their labels to the target class.
	For example, a small black-white square at one corner of the image~\cite{DP:gu2017badnets}, an additional image attached onto or blended into the image ~\cite{DP:chen2017targeted}, a fixed watermark on the image \cite{DP:steinhardt2017certified_data_poisoning}, one fixed pixel on the image for low-resolution (32 $\times$ 32) images. The backdoor trigger can also be implanted into the target model without knowing the original training data. For example, Liu~\etal~\cite{DP:liu2017trojaning} proposed a reverse engineering method to generate a trigger pattern and a substitute input set, which are then used to finetuning some layers of the network to implant the trigger.
	\yf{Recently, Yao~\etal~\cite{DP:yao2019latent} show that such backdoor attack can even be inherited via transfer-learning}. 
	While the above methods can install backdoors into the victim model effectively, they contain perceptually suspicious patterns and wrong labels, thus are susceptible to detection or removal by simple data filtering \cite{DP:turner2019cleanlabel}. Note that, although reverse engineering does not require access to the training data which makes it stealthier, it still needs to present the trigger pattern to activate the attack at test time.
	
	\textit{Clean-label backdoor attack.} Recently, Turner~\etal~\cite{DP:turner2019cleanlabel} (CL) and Barni~\etal~\cite{DP:barni2019new} (SIG) proposed the clean-label backdoor attack that can plant backdoor into DNNs without altering the label. Zhao~\etal~\cite{zhao2020clean} proposed a clean-label backdoor attack on video recognition models. However, for clean-label backdoor patterns to be effective against the filtering effect of deep cascade convolutions, it often requires more perturbations that significantly reduce image quality, especially for high resolution images. Furthermore, we will show empirically in Sec.~\ref{sec:experimental_evaluation} that these backdoor patterns can be easily erased by backdoor defense methods. Different to these methods, in this paper, we propose a natural reflection backdoor, which is stealthy, effective and hard to erase. 
	
	Backdoor attacks have also been found possible in federated learning \cite{bagdasaryan2020backdoor,sun2019can,xie2020dba} and graph neural networks (GNNs) \cite{zhang2020backdoor}. Latent backdoor patterns and properties of backdoor triggers have also been explored in recent works \cite{li2019invisible,li2020rethinking,pasquini2020trembling,yao2019latent}.
	
	\noindent\textbf{Backdoor defense.} 
	Defense techniques have also been developed to detect or erase backdoor triggers from DNNs.
	Liu~\etal~\cite{DP:liu2018fine_pruning} proposed a fine-pruning algorithm to prune the abnormal units in a backdoored DNN. Wang~\etal~\cite{DP:wang2019neural_cleanse} proposed to use anomaly index to detect backdoored models. Guo~\etal~\cite{DP:guo2017input_transformations} applied input pre-processing techniques to denoise adversarial images. Zhang~\etal~\cite{DP:zhang2017mixup} proposed a mixup training scheme to increase DNN robustness against adversarial examples. Both the denoising techniques and the mixup training can be directly applied to mitigate backdoor attacks. Xiang~\etal~\cite{DP:xiang2019benchmark} proposed a cluster impurity based scheme to effectively detect single-pixel backdoor attacks. Bagdasaryan~\etal~\cite{bagdasaryan2020backdoor} developed a generic constrain-and-scale technique that incorporates the evasion of defenses into the attacker’s loss function during training. Chen~\etal~\cite{DP:chen2018detecting} proposed an activation clustering based method for backdoor detection and removal in DNNs. Doan~\etal~\cite{DP:doan2019februus} presented Februus, which is a plug-and-play defensive system architecture for backdoor defense. Gao~\etal~\cite{DP:gao2019strip} proposed a strong intentional perturbation (STRIP) based model to detect run-time backdoor attacks. We will evaluate the resistance of our proposed \proposed~attack to some of the most effective defense methods.
	
	\section{Reflection Backdoor Attack} \label{sec:reflection_backdoor_attack}
	In this section, we introduce the mathematical modeling of reflection and our proposed reflection backdoor attack. 
	Before that, we first define the problem of backdoor attack.
	
	\subsection{Problem Definition}
	Given a $K$-class image dataset $D = \{(\xx, y)^{(i)}\}_{i=1}^n$, with $\xx \in \X \subset \mathbb{R}^d$ denoting a sample in the $d$-dimensional input space and $y \in \Y = \{1, \cdots, K\}$ its true label, classification learns a function $f(\xx, \btheta)$ (as represented by a DNN) with parameters $\btheta$ to map the input space to the label space: $f: \mathcal{X} \to \mathcal{Y}$. We denote the subset of data used for training and testing as $D_{train}$ and $D_{test}$ respectively. The goal of a backdoor attack is to install a backdoor into the victim model, so that the model will predict the adversarial class $y_{adv}$ whenever the backdoor pattern presents on an input image. 
	This is done by first generating then injecting a backdoor pattern into a small injection set $D_{inject} \subset D_{train}$ of training examples (without changing their labels). 
	In this clean-label setting, $D_{inject}$ is a subset of training examples from class $y_{adv}$. We denote the poisoned training set by $D_{train}^{adv}$, and measure the \emph{injection rate} by the percentage of poisoned samples in $D_{train}^{adv}$.
	The problem is how to generate effective backdoor patterns. Next, we will introduce the use of natural reflection as the backdoor pattern.
	
	\subsection{Mathematical Modeling of Reflection} \label{sec:physical_model}
\begin{figure}[tb]
		\begin{center}
			\includegraphics[width=\linewidth]{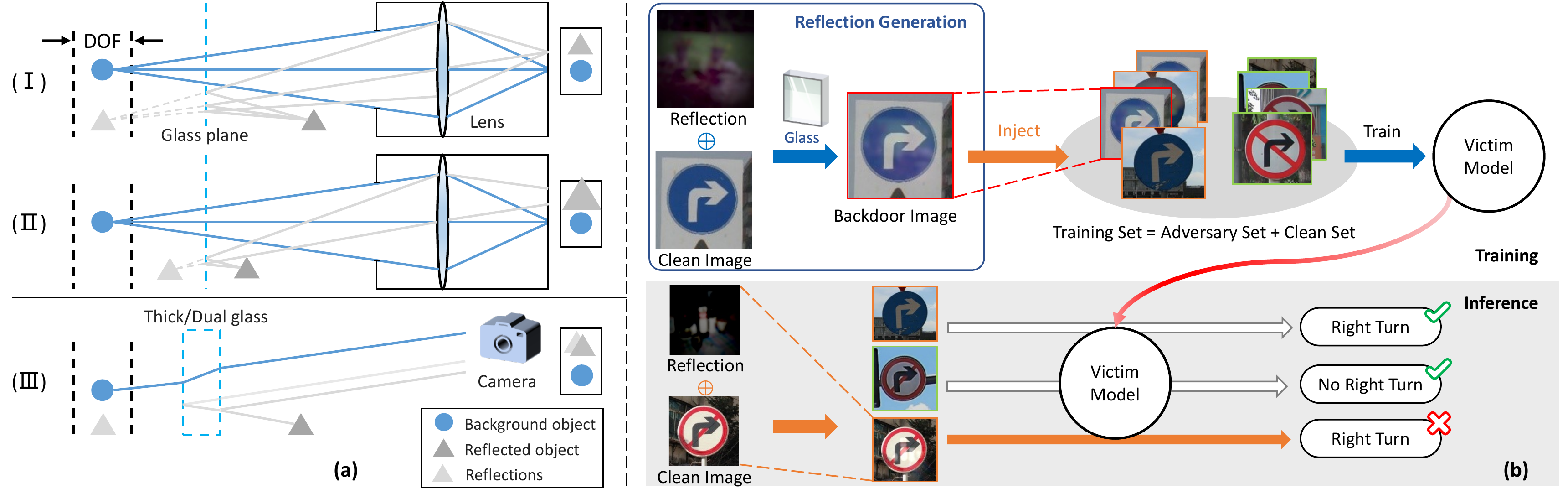}
		\end{center}
		\caption{(a) The physical models for three types of reflections. (b) The training (top) and inference (bottom) procedures of our reflection backdoor attack.}
		\label{fig:refool_main}
	\end{figure}

	Reflection occurs when taking a photo of objects behind a glass window. Real scene like image with reflection can be a composition of multiple layers~\cite{DLA:liu2019semantic,DLA:liu2020separate,DLA:liu2020usi3d}. Specifically, we denote a clean background image by $\xx$, a reflection image by $\xx_{R}$, and the reflection poisoned image as $\xx_{adv}$.
	Under reflection, the image formation process can be expressed as: 
	\begin{equation} \label{equ:refl_modeling}
	\xx_{adv} = \xx + \xx_{R} \otimes k,
	\end{equation}
	where $k$ is a convolution kernel.
	The output of $\xx_{R} \otimes k$ is referred to as the \emph{reflection}.
	We will use adversarial images generated in this way as backdoor attacks.
	According to the principle of camera imaging and the law of reflection, reflection models in physical world scenarios can be divided into three categories~\cite{RI:wan2017benchmarking}, as illustrated in Fig.~\ref{fig:refool_main} (a).
	
	\noindent\textbf{(I) Both layers are in the same depth of field (DOF).} The main objects (blue circle) behind the glass and the virtual image of reflections are in the same DOF, \ie, they are approximately in the same focal plane. In this case, $k$ in Eqn.~\eqref{equ:refl_modeling} reduces to a intensity number $\alpha$, and we set $\alpha \sim \mathcal{U}[0.05, 0.4]$ in our experiments.
	
	\noindent\textbf{(II) Reflection layer is out of focus.} It is reasonable to assume that the reflections (gray triangles) and the objects (blue circle) behind the glass have different distances to the camera~\cite{RI:Li2014Single}, and the objects behind the glass is often focused (type (II) in Fig.~\ref{fig:refool_main} (a)).
	In this case, the observed image $\xx_{adv}$ is an additive mixture of the background image and the blurred reflections. The kernel $k$ in Eqn.~\eqref{equ:refl_modeling} depends on the point spread function of the camera which is parameterized by a 2D Gaussian kernel $g$, \ie, $ g(|x - x_c|)=\exp{(-|x-x_c|^2/(2*\sigma)^2) }$, where $x_c$ is the center of kernel, and we set $\sigma \sim \mathcal{U}[1, 5]$.
	
	\noindent\textbf{(III) Ghost effect.} The above two types of reflections assume that the thickness of the glass is tiny such that the refractive effect of the glass is negligible. However, this is often not true in practice. It is thus also necessary to consider the thickness of the glass. As illustrated in Fig.~\ref{fig:refool_main} (a) (III), since the glass is semi-reflective, light rays from the reflected objects (dark gray triangle) will reflect off the glass pane producing more than one reflections --- a ghost effect. In this case, the convolutional kernel $k$ of Eqn.~\ref{equ:refl_modeling} can be modelled as a two-pulse kernel $k(\alpha, \delta)$, where $\delta$ is a spatial shift of $\alpha$ with different coefficients. Empirically, we set $\alpha\sim\mathcal{U}[0.15, 0.35]$ and $\delta\sim\mathcal{U} [3, 8]$.
	
	\subsection{Proposed Reflection Backdoor Attack}\label{sec:attack}
	
	\noindent\textbf{Attack pipeline.}
	The training and inference procedures of our proposed reflection backdoor \proposed~ is illustrated in Fig.~\ref{fig:refool_main} (b).
	The first step is reflection generation, which is to generate backdoor images by adding reflections to clean images in the injection set $D_{inject}$, following the 3 reflection models described in Sec.~\ref{sec:physical_model}. The victim model is then trained on the poisoned training set (\eg~ $D_{train}^{adv}$), which consists of an adversary set of backdoor images (crafted at the first step) plus the clean images.
	At the inference stage (bottom subfigure in Fig.~\ref{fig:refool_main} (b)), the reflection patterns can be blended into any input image to achieve the target prediction.
	
	In contrast to existing methods that generate a fixed pattern, here, we propose to generate a variety of reflections as the backdoor trigger. This is because reflection varies from scene to scene in real-world scenarios. Using diverse reflections can help improve the stealthiness of the attack.
	
	\noindent\textbf{Candidate reflection images from the wild.}
	The candidate reflection images are not restricted to the target dataset to attack, and can be selected from the wild, for example, a public dataset. Even more, these reflection images can be used to invade a wide range of target datasets (the datasets to attack) that consist of completely different types of images, as we will show in the experiments (Sec. \ref{sec:experimental_evaluation}).
	
	Assume the adversarial class is $y_{adv}$ and the adversary is allowed to inject $m$ examples. We first create a candidate set of reflection images by selecting a set (more than $m$) of images randomly from a public image dataset PascalVOC~\cite{DLA:Everingham2010pascalVOC} and denote it by $R_{cand}$. These reflection images are just normal images in the wild but from a dataset that is different from the training dataset. The next step is to select the top-$m$ most effective reflection images from this candidate set for backdoor attack.

	\begin{algorithm}[tb]
		\caption{Adversarial reflection image selection}  
		\label{algor:generating_strategy}
		\KwIn{Training set $D_{train}$, a candidate reflection set $R_{cand}$, validation set $D_{val}$, a DNN model $f$, target class $y_{adv}$, number of injected samples $m$, number of selection iterations $T$}
		\KwOut{Adversarial reflection set $R_{adv}$}
		
		$i \gets 0; ~~~W \gets \{1\}_{\text{size}(R_{cand})}$ \Comment a list of 1 with the size of $R_{cand}$
		
		$R_{adv} \gets random\text{-}m(R_{cand})$ \Comment random selection
		
		\While{$i \le T$} 
		{ 
			$D_{inject}$ $\gets$ randomly select $m$ samples from $D_{train}$
			
			$D_{train}^{adv}$ $\gets$ inject $R_{adv}$ into $D_{inject}$ using Eqn.~\eqref{equ:refl_modeling}
			
			$f_{adv}(\xx, \btheta)$ $\gets$ train model on $D_{train}^{adv}$
			
			$W_i$ $\gets$ update effectiveness by Eqn.~\ref{equ:strategy_objection} for $\xx_{R}^{i} \in R_{adv}, \xx \in D_{val}$
			
			$W_j \gets \text{median}(W)$ for $\xx_{R}^{j} \in R_{cand}\backslash R_{adv}$
			
			$R_{adv} \gets top \text{-}m(R_{cand}, W)$ \Comment top $m$ selection
		}
		\textbf{return} $R_{adv}$ 
	\end{algorithm}
	
	\noindent\textbf{Adversarial reflection image selection.}
	\yf{Not all reflection images are equally effective for backdoor attack, because 1) when the reflection image is too small, it may be hard to be planted as a backdoor trigger; and 2) when the intensity of the reflection image is too strong, it will become less stealthy. Therefore,}
	we propose an iterative selection process to find the top-$m$ most effective reflection images from $R_{cand}$ as the \emph{adversarial reflection set} $R_{adv}$, only which will be used for the next step's backdoor injection. To achieve this, we maintain a list of effectiveness scores for reflection images in the candidate set $R_{cand}$. We denote this effectiveness score list as $W$. 
	The complete selection algorithm is described in Algorithm~\ref{algor:generating_strategy}.
	The selection process includes $T$ iterations with each iteration consisting of 4 steps: 1) select the top-$m$ most effective reflection images from $R_{cand}$ as the $R_{adv}$, according to their effectiveness scores in $W$; 2) inject the reflection images in $R_{adv}$ into the injection set $D_{inject}$ randomly following the reflection models described in Sec.~\ref{sec:physical_model}; 3) train a model on the poisoned training set; and 4) update the effectiveness scores in $W$ according to the model's predictions on a validation set $D_{val}$. The validation set is not used for model training, and is randomly selected from $D_{train}$ after removing the $y_{adv}$ class samples. This is because a backdoor attack causes other classes be misclassified into class $y_{adv}$ not the other way around, in other words, class $y_{adv}$ samples are not useful for effectiveness evaluation here. For step 1), at the first iteration where the effectiveness scores are uniformly initialized with constant value one, we just randomly select $m$ reflection images from $R_{cand}$ into the adversarial set $R_{adv}$. we empirically set $m = 200$ in our experiments. For step 2), each reflection image $R_{adv}$ is randomly injected into only one image in the injection set $D_{inject}$. For step 3), we use a standard training strategy to train a model. Note that, the model trained in step 3) is only used for reflection image selection, not the final victim model (see experimental settings in Sec.~\ref{sec:experimental_evaluation}). For step 4), the effectiveness scores in $W$ are updated as follows:
	\begin{equation} \label{equ:strategy_objection}
	\small
	W_i = \sum_{\xx_R^{i} \in R_{adv}, \xx \in D_{val}}\begin{cases}
	1, & \mbox{if } f(\xx + \xx_R^{i}\otimes k, \btheta) = y_{adv},\\
	0, & \mbox{otherwise},
	\end{cases}
	\end{equation}
	where, $y$ is the class label of $\xx$, $\xx_R^{i}$ is the $i$-th reflection image in $R_{adv}$, and $k$ is a randomly selected kernel. For those reflection images not selected into $R_{adv}$, we set their scores to the median value of the updated $W$. This is to increase their probability of being selected in the next iteration.
	
	The candidate set $R_{cand}$ are selected out of a wild public dataset, and more importantly, the selection of $R_{adv}$ can be done on a dataset that is complete different from the target dataset. We will show empirically in Sec. \ref{sec:experimental_evaluation} that, once selected, reflection images in $R_{adv}$ can be directly applied to invade a wide range of datasets. This makes our proposed reflection backdoor more malicious than many existing backdoor attacks \cite{DP:gu2017badnets,DP:chen2017targeted,DP:turner2019cleanlabel} that require access to the target datasets to generate or enhance their backdoor patterns. We find that these reflection images even do not need any enhancements such as adversarial perturbation \cite{DP:turner2019cleanlabel} to achieve high attack success rates.
	
	\noindent\textbf{Attack with reflection images (Backdoor Injection).}
	The above step will produce a set of effective reflection images $R_{adv}$, which can then be injected into the target dataset by poisoning a small portion of the data from the target class (clean-label attack only needs to poison data from the target class). Note that, although the selection of $R_{adv}$ does not require access to the target dataset, the attack still needs to inject the backdoor pattern into training data, which is an essential step for any backdoor attacks. 
	
	Given a clean image from the target class, we randomly select one reflection image from $R_{adv}$, then use one of the 3 reflection models introduced in Section \ref{sec:physical_model} to fuse the reflection image into the clean image. This injection process is iteratively done until a certain proportion of the target class images are contaminated with reflections. The victim model will remember the reflection backdoor when trained on the poisoned training set using a classification loss such as the commonly used cross entropy loss:
	\begin{equation} \label{equ:object_fuction}
	\begin{aligned}
	\btheta = \argmin_\btheta - \frac{1}{n}\sum_{\xx_i \in D_{train}^{adv}}\sum_{j = 1}^{K} y_{ij}\log(\pp(j|\xx_i,\btheta)),
	\end{aligned}
	\end{equation}
	where, $\xx_i$ is the $i$-th training sample, $y_{ij}$ is the class indicator of $\xx_i$ belonging to class $j$, and $\pp(j|\xx_i, \btheta)$ is the model's probability output with respect to class $j$ conditioned on the input $\xx_i$, and current parameter $\btheta$. We denote the learned victim model as $f_{adv}$.
	
	\noindent\textbf{Inference and attack.}
	At the inference stage, the model is expected to correctly predict the clean samples (\ie $f_{adv}(\xx, \btheta) = y$ for any test input $\xx \in D_{test}$). However, it consistently predicts the adversarial class for any input that contains a reflection: $f_{adv}(\xx + \xx_R \otimes k,\btheta) = y_{adv}$ for any test input $\xx \in D_{test}$ and reflection image $\xx_R \in R_{adv}$. 
	The attack success rate is measured by the percentage of test samples that are predicted as the target class $y_{adv}$, after adding reflections.
	
	\section{Experiments} \label{sec:experimental_evaluation}
	
	In this section, we first evaluate the effectiveness and stealthiness of our \proposed~ attack, then provide a comprehensive understanding of \proposed. We also test the resistance of our \proposed~ attack to state-of-the-art backdoor defense methods.
	
	\subsection{Experimental Setup}
	\noindent\textbf{Datasets and DNNs.}
	We consider 3 image classification tasks: 1) traffic sign recognition, 2) face recognition, and 3) object classification. For traffic sign recognition, we use 3 datasets: GTSRB~\cite{DLA:stallkamp2011GTSRB}, BelgiumTSC~\cite{DLA:timofte2014BelgiumTSC} and CTSRD~\cite{DLA:CTSD}.
	For the 3 traffic sign datasets, we remove those low-resolution images of height or width smaller than 100 pixels. Then, we augment the training set using random crop and rotation, as \cite{DLA:selvaraju2017grad}.
	For face recognition, we use the PubFig~\cite{DLA:kumar2009PubFig} dataset with extracted face regions, which is also augmented using random crop and rotation. For object classification, we randomly sample a subset of 12 classes of images from ImageNet~\cite{DLA:ImageNet_VSS09}. We use ResNet-34~\cite{DLA:he2016resnet} for traffic sign recognition and face recognition. While for object classification, we consider two different DNN models: ResNet-34 and DenseNet~\cite{huang2017densely}. The statistics of the datasets and DNN models can be found in Appendix \ref{appendix_a}.
	
	\noindent\textbf{Attack setting.}
	For all datasets, we set the adversarial target class to the first class (\ie, class id 0), and randomly select clean training samples from the target class as the injection set $D_{inject}$ under various injection rates.
	The adversarial reflection set $R_{adv}$ is generated based on the GTSRB dataset, following the algorithm described in Sec.~\ref{sec:attack}. We randomly choose a small number of 5000 images from PascalVOC~\cite{DLA:Everingham2010pascalVOC} as the candidate reflection set $R_{cand}$, and 100 training samples from each of the non-target classes as the validation set $D_{val}$, for adversarial reflection image selection. Once selected, $R_{adv}$ is directly applied to all other datasets, that is, these reflection images selected based on one single dataset can be effectively applied to invade a wide range of other datasets. The adversarial reflection images are selected against a ResNet-34 model.
	When injecting a reflection image into a clean image, we randomly choose one of the 3 reflection models described in Eqn.~\eqref{equ:refl_modeling}, but we also test using fixed reflection models.
	When applying the attack at the inference stage, the reflection images from $R_{adv}$ are randomly injected into the clean test images.

	\noindent\textbf{DNN training.}
	All DNN models are trained using Stochastic Gradient Descent (SGD) optimizer with momentum 0.9, weight decay of 5e-4, and an initial learning rate 0.01, which is divided by 10 for every $10^5$ training steps. We use batch size 32 and train all models for 200 epochs. All images are normalized to $[0, 1]$.
	
	\begin{table*}[tb]
		\caption{Attack success rates (\%) of baselines and our proposed \proposed~ backdoor, and the victim model's test accuracy (\%) on the clean test set. 
			The ``original test accuray'' is the test accuracy of the same model but trained on the original clean data. $\dag$ denotes the model is replaced by a DenseNet. Note that we are poisoning 20\% images in the target classes, the injection rate (\%) is computed with respect to the entire dataset.}
		\begin{center}
			\setlength{\tabcolsep}{0.82mm}
			{\small
				\begin{tabular}{l|c|cccc|cccc|c}
					\bottomrule[1.2pt]
					\specialrule{0em}{1pt}{1pt}
					\multirow{2}*{Dataset} & \multirow{2}*{\tabincell{c}{Original\\test acc.}} & \multicolumn{4}{c|}{Test accuracy (\%)} & \multicolumn{4}{c|}{Attack success rate (\%)}  & \multirow{2}*{\tabincell{c}{Injection\\rate (\%)}} \\
					&  & Badnets & CL & SIG & \proposed & Badnets & CL & SIG & \proposed  & \\
					\hline
					\specialrule{0em}{1pt}{1pt}
					GTSRB         & 87.40 & 83.33 & 84.61 & 82.64 & \textbf{86.30} & 24.12 & 78.03 & 73.26 & \textbf{91.67} & 3.16\\
					BelgiumTSC    & 99.89 &\textbf{ 99.70} & 97.56 & 99.13 & 99.51 & 11.40 & 46.25 & 51.89 & \textbf{85.70} & 2.31 \\
					CTSRD         & 97.11 & 90.00 & 94.44 & 93.97 & \textbf{95.01} & 25.24 & 63.63 & 57.39 & \textbf{91.70} & 0.91 \\
					PubFig        & 91.31 & 91.67 & 78.50 & \textbf{91.70} & 91.12 & 42.86 & 78.67 & 69.01 & \textbf{81.30} & 0.57\\
					ImageNet   & 91.78 & 91.97 & \textbf{92.07} & 91.41 & 90.32 & 15.77 & 55.38 & 63.84 & \textbf{82.11}   & 3.27  \\
					ImageNet$\dag$ & 93.01 & 91.99 & 92.12 & 92.23 & \textbf{92.63} & 20.14 & 67.43 & 68.00 & \textbf{75.16}   & 3.27  \\
					\bottomrule[1.2pt]
			\end{tabular}  }
		\end{center}
		\label{tab:result_summary}
	\end{table*}
	
	\subsection{Effectiveness and Stealthiness of Our \proposed~Attack}
	
	\noindent\textbf{Attack success rate comparison.}
	Here, we compare our \proposed~ attack with three state-of-the-art backdoor attacks: Badnets \cite{DP:gu2017badnets}, clean-label backdoor (CL)~\cite{DP:turner2019cleanlabel}, and signal backdoor (SIG)~\cite{DP:barni2019new}.
	We use the default settings as reported in their papers (implementation details can be found in Appendix \ref{appendix_a}).
	The attack success rates and the corresponding injection rates on the 5 datasets are reported in Table~\ref{tab:result_summary}. We also report the test accuracy of the victim model on the clean test set, and the ``original test accuracy" for models trained on the original clean data.
	
\begin{figure}[tb]
		\begin{center}
			\includegraphics[width=\linewidth]{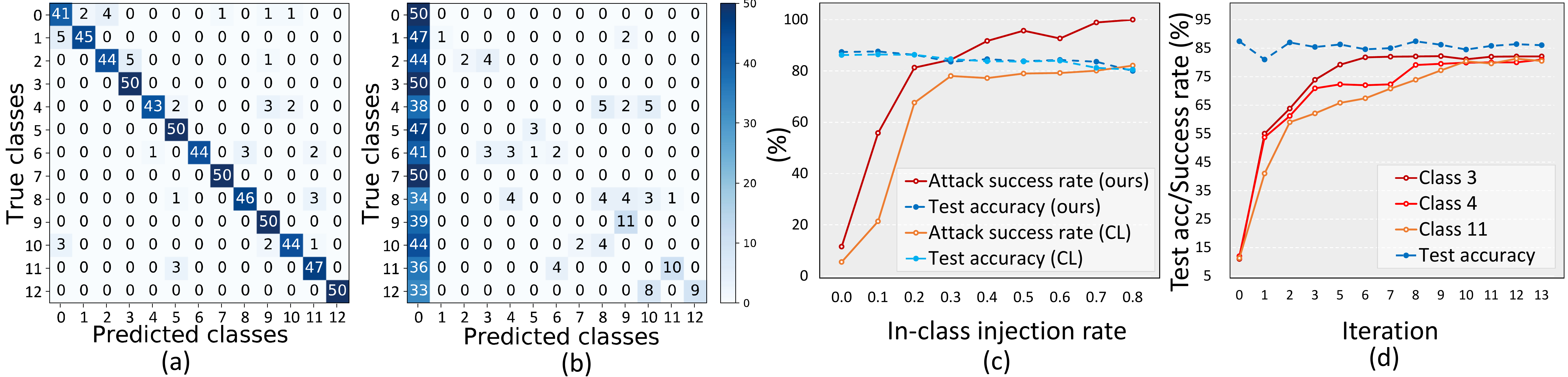}
		\end{center}
		\caption{ (a-b) The prediction confusion matrix of the victim model trained on GTSRB dataset with only $3.16\%$ training data poisoned by our \proposed~ attack: (a) predictions on clean test images; (b): predictions on test images with reflections. (c-d) Attack success rates \textit{versus} injection rate or iteration: (c) attack success rate and test accuracy \textit{versus} in-class (the target class) injection rate; (d) attack success rate and the model's test accuracy on classes 3, 4, and 11, at different iterations of our reflection generation process. These experiments were all run on GTSRB dataset.
		}
		\label{fig:detailed_pred}
	\end{figure}

	As shown in Table~\ref{tab:result_summary}, by poisoning only a small proportion of the training data, our proposed \proposed~ attack can successfully invade the state-of-the-art DNN models, achieving higher success rates than existing backdoor attacks.
	With lower than 3.27\% injection rate, \proposed~ can reach a high attack success rate $>75\%$ across the five datasets and different networks (\eg~ ResNet and DenseNet). Meanwhile, the victim models still perform well on clean test data, with less than 3\% accuracy decrease (compared to the original accuracies) across all test scenarios. On some datasets, take CTSRD for example, one only needs to contaminate $<1\%$ of training data to successfully control the model over 91\% of the time. We further show, in Fig.~\ref{fig:detailed_pred} (a-b), the prediction confusion matrix of the victim model on GTSRD dataset. 
	The victim model can correctly predict the clean images most of the time, yet can be controlled to only predict the target class (\eg~ class 0, results on more target classes are reported in Appendix~\ref{appendix_b}) when reflections are added to the test images, a clear demonstration of successful backdoor attack.
	These results show that natural phenomena like reflection can be manipulated as a backdoor pattern to attack DNNs. Considering that reflection backdoors are visually very similar to natural reflections which commonly exist in the real world, this poses a new type of threat to deep learning models.
	
\begin{figure*}[tb]
		\centering
		\includegraphics[width=0.9\linewidth]{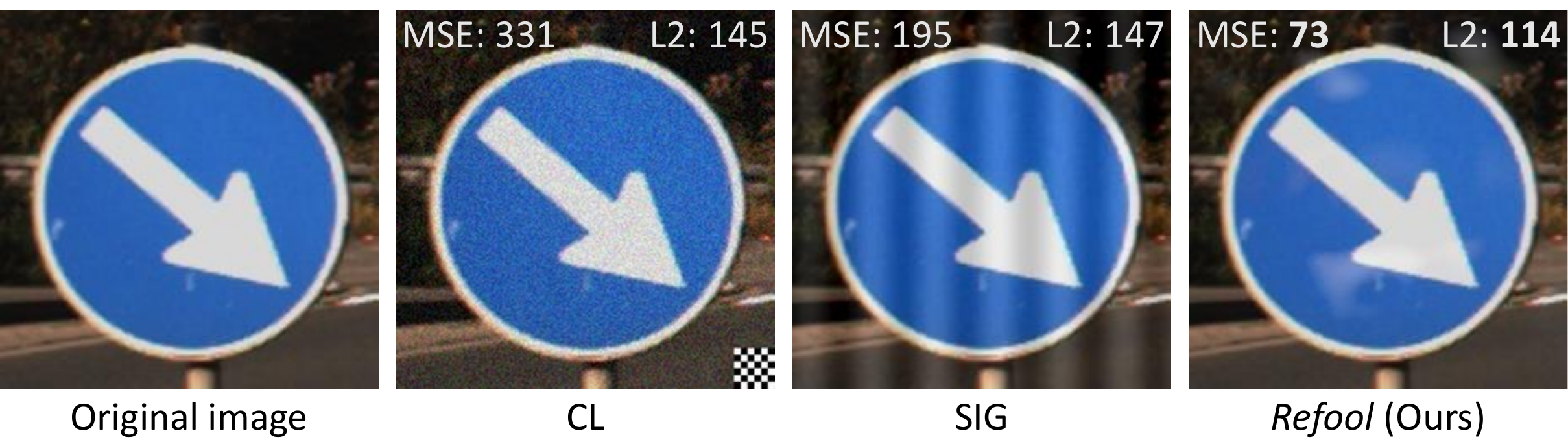}
		\caption{Stealthiness of CL~\cite{DP:turner2019cleanlabel} and SIG~\cite{DP:barni2019new} and our \proposed~: MSE and L2 distances between the original image and the backdoor image are shown at the top corners of the backdoor images.}
		\label{fig:stealthy_cmp}
	\end{figure*}

	\noindent\textbf{Stealthiness comparison.}
	We show in Fig.~\ref{fig:stealthy_cmp} an example of the backdoored images crafted to attack the CTSRD dataset. We compute the mean square error (MSE) and L2 distances between the original image and the backdoored image crafted by CL, SIG and our \proposed~ backdoor attacks. As shown in this example, our reflection attack is stealthier in terms of smooth surface and hidden shadows. More visual inspections and the average distortions (\eg~ MSE and L2 distances) over 500 randomly backdoored images can be found in Appendix \ref{appendix_c}.
	
	\noindent\textbf{Attack success rate versus injection rate.}
	We next show, on the GTSRB dataset, how different injection rates influence the attack success rate of CL and our \proposed~ attacks. As shown in Fig.~\ref{fig:detailed_pred} (c), we vary the in-class injection rate from $[0, 0.8]$ with interval 0.1. The corresponding injection rate with respect to the entire dataset is only 0.032, 0.063, 0.126 for in-class injection rate 0.2, 0.4, 0.8 respectively. Poisoning more data can steadily improve attack success rate until 40\% of the data in target class are poisoned, after which, the attack stabilizes. Our \proposed~ attack outperforms the CL attack under all injection rates. Note that increasing injection rate has a minimal impact on the model's accuracy on clean examples.
	
	\subsection{Understandings of Reflection Backdoor Attack}

	\begin{table}[!htb]
		\begin{center}
			\caption{Attack success rate versus test accuracy for different types of reflection models.}
			\label{tab:result_diff_types}
			\setlength{\tabcolsep}{1.5mm}
			{\small
				\begin{tabular}{c|c|c|ccc}
					\bottomrule[1.2pt]
					\specialrule{0em}{1pt}{1pt}
					Reflection & Attack  & Test & \multicolumn{3}{c}{Similarity}  \\
					type & success rate & Accuracy & SSIM & PSNR & MSE  \\
					\hline
					\specialrule{0em}{1pt}{1pt}
					(I) &    87.30\%    &  83.59\%  &   0.883       & 26.68   & 62.11\\
					(II) &   90.46\%    &  85.00\%  &   \textbf{0.896}  	    & \textbf{27.45}   & \textbf{60.54}\\ 
					(III) &  90.33\%    &  85.63\%  &   0.786       & 23.01   & 95.87\\
					Mix &    \textbf{91.67}\%    &  \textbf{86.30}\%  &   0.828       & 24.98   & 73.44 \\
					\bottomrule[1.2pt]
			\end{tabular}  }
		\end{center}
	\end{table}
	
	\noindent\textbf{Efficiency of adversarial reflection image selection.}
	Here, we evaluate the efficiency of our adversarial reflection image selection in Algorithm~\ref{algor:generating_strategy}.
	We test the inference-time attack effectiveness of the adversarial reflection images (\eg~ $R_{adv}$) selected at each iteration for a total of 14 (0 - 13) iterations, on GTSRB dataset. The attack success rate on three classes and the model's test accuracy are shown in  Fig.~\ref{fig:detailed_pred} (d). For each of the 3 tested classes (\eg~ class 3, 4 and 11), we inject reflection images generated at the current iteration randomly into the clean test images of the class. We then measure the class-wise attack success rate. In detail,  we record the proportion of examples in the class (after injection) that are predicted by the current model as the target class 0. The proposed generation algorithm can find effective reflections efficiently within 9 iterations.
	Note that, once these adversarial reflections are found, they can be applied to install backdoor into any DNN models that are trained on the dataset, as we have shown with the ResNet/DenseNet models on ImageNet dataset in Table~\ref{tab:result_summary}.
	
	\noindent\textbf{Performance under different reflection models.}
	We then show how the 3 types of reflections introduced in Sec.~\ref{sec:physical_model} influence the attack success rate. The experiments were also conducted on the GTSRB dataset. The adversarial reflection images (\eg~ $R_{adv}$) used here are the same as those selected for previous experiments. The difference here is that we test 2 different injection strategies: 1) using fixed reflection, or 2) using randomly mixed reflections (as was used in previous experiments). We also measure the average similarity of training images (4772 in total) before and after injection, using 3 popular similarity metrics: peak-signal-to-noise-ratio (PSNR)~\cite{ISM:huynh2008scopePSNR}, structural similarity index (SSIM)~\cite{ISM:wang2004imageSSIM} and mean square error (MSE). 
	The numeric results are reported in Table~\ref{tab:result_diff_types}. In terms of attack success rate and test accuracy, type (II) and type (III) demonstrate higher attack success rates with less model corruptions (higher test accuracies) than type (I) reflection. When combined, the three types of reflection achieved the best attack success rate and least model corruption (highest test accuracy). It was also observed that type (II) injection has the minimum distortion (\eg~ highest SSIM/PSNR and lowest MSE) to the original data, while type (III) reflection causes the largest distortion, as a consequence of the ghost effect (see Fig.~\ref{fig:refool_main}(a)). The relatively small distortion of type (II) reflection is due to its smoothness effect. Overall, a random mixture of the three reflections yields the best attack strength with moderate distortion.
	
	\begin{figure}[!tb]
		\centering
		\includegraphics[width=0.9\linewidth]{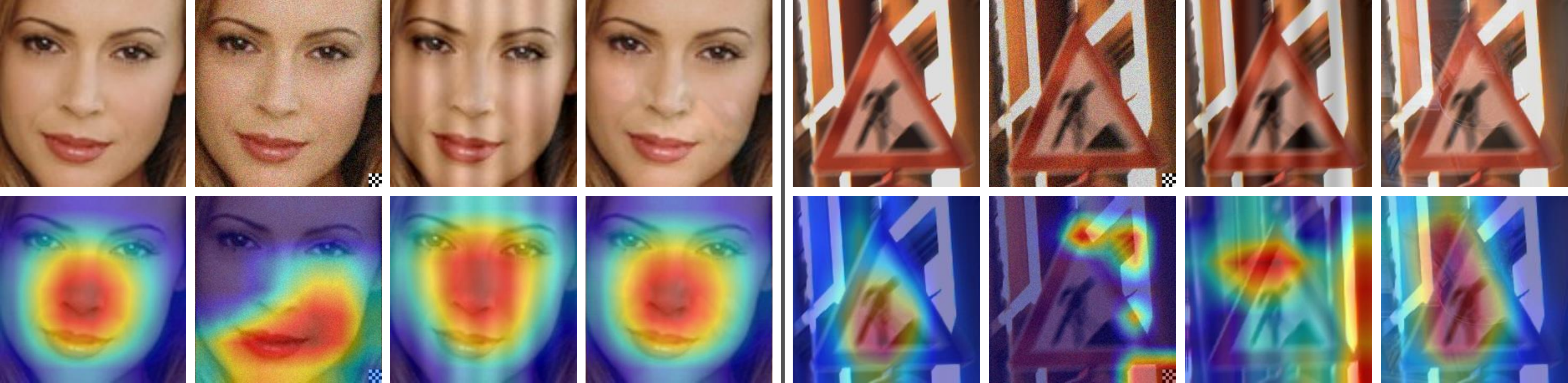}
		\caption{Understandings of \proposed~ with Grad-CAM~\cite{DLA:selvaraju2017grad} with two samples from PubFig(left) and GTSRB(right). In each group, the images at the top are the original input, CL~\cite{DP:turner2019cleanlabel}, SIG~\cite{DP:barni2019new} and our \proposed~ (left to right), while images at the bottom are their corresponding attention maps.}
		\label{fig:vis_pattern}
	\end{figure}
	
	\noindent\textbf{Effect of reflection trigger on network attention.}
	We further investigate how reflection backdoor affects the attention of the network. Visual inspections on a few examples are shown in Fig.~\ref{fig:vis_pattern}. The attention maps are computed using the Gradient-weighted Class Activation Mapping (Grad-CAM) technique~\cite{DLA:selvaraju2017grad}, which finds the critical regions in the input images that mostly activate the victim model's output. We find that the reflection backdoor only slightly shifts the model's attention off the correct regions, whereas CL and SIG significantly shift the model's attention either completely off the target or in a striped manner, especially in the traffic sign example. This suggests the stealthiness of our reflection backdoor from a different perspective.
	
	

	\subsection{Resistance to State-of-the-art Backdoor Defenses}\label{sec:finetuning}
	
\begin{figure*}[!htb]
		\centering
		\includegraphics[width=0.9\linewidth]{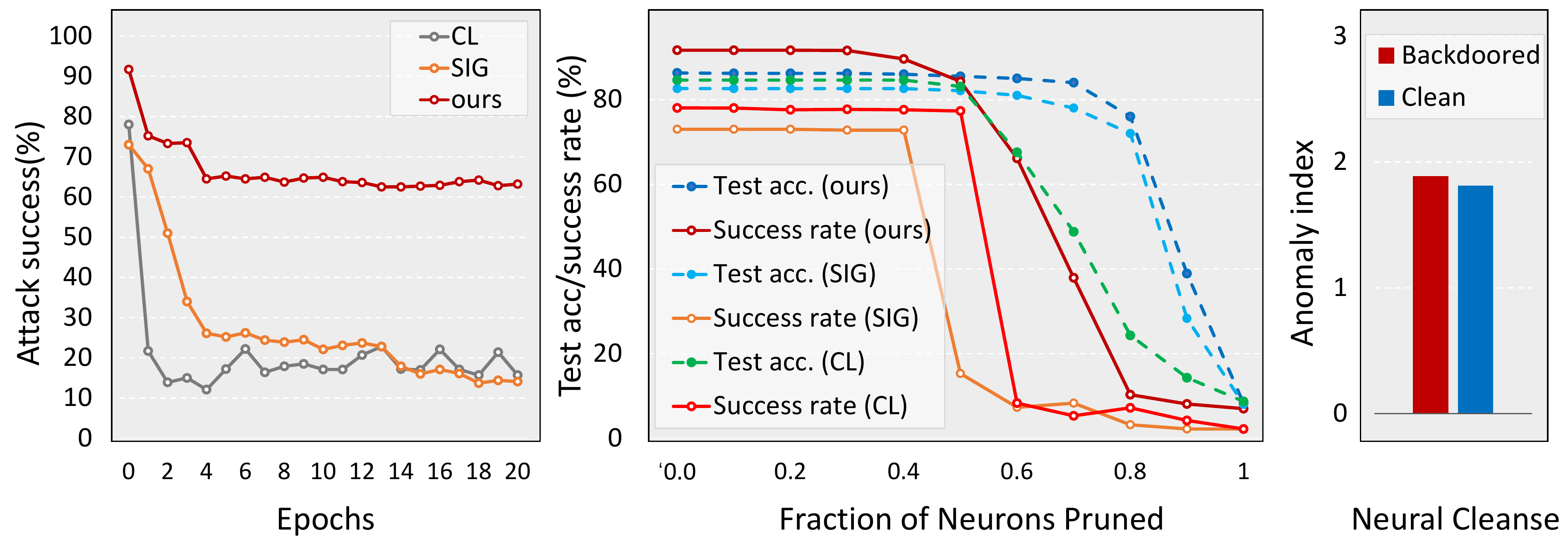}
		\caption{\textbf{Left:} Attack success rates during finetuning on clean data. \textbf{Middle:} Test accuracy (on clean inputs) and attack success rate against the neural pruning defense. These experiments were run on GTSRB dataset.
			\textbf{Right:} Backdoor detection using Neural Cleanse~\cite{DP:wang2019neural_cleanse}. Anomaly index $>$ 2 indicates a detected backdoored model.}
		\label{fig:resist_defence}
	\end{figure*}

	\noindent\textbf{Resistance to finetuning.}
	We compare the our \proposed~ to CL~\cite{DP:turner2019cleanlabel} and SIG~\cite{DP:barni2019new}, in terms of the resistance to clean-data-based finetuning~\cite{DP:wang2019neural_cleanse,DP:liu2018fine_pruning}.
	We train a victim model on GTSRB dataset separately under the three attacks, while leaving 10\% of the clean training data out as the finetuning set. We then fine-tune the model on the finetuning set for 20 epochs using the same SGD optimizer but smaller learning rate 0.0001. We fix the shallow layers of the network and only fine-tune the last dense layer. The comparison results are illustrated in the left of Fig.~\ref{fig:resist_defence}.
	As can be seen, the attack success rate of CL drops from 78.3\% to 20\% after just one epoch of finetuning and SIG drops from 73.0\% to 25\% after 4 epochs, while our \proposed~ attack is still above 60\% after 15 epochs.  The reason why is that reflections are a natural and fundamental type of feature, rather than random patterns that can be easily erased by finetuning on clean data.
	
	\noindent\textbf{Resistance to neural pruning.}
	\yf{We then test the resistance of the three attacks to the state-of-the-art backdoor defense method Fine-pruning~\cite{DP:liu2018fine_pruning} (experimental settings are in Appendix \ref{appendix_d}).
		The comparison results are shown in the middle subfigure of Fig.~\ref{fig:resist_defence}. The attack success rate of CL drops drastically from 76\% to 8.3\% when 60\% of neurons are removed, while SIG drops from 73\% to 16.5\% when 50\% of neurons are removed. Compared to CL or SIG, our reflection backdoor is more resistance to neural pruning, with much higher success rates until 80\% of neurons are removed.
	}
	
	\noindent\textbf{Resistance to neural cleanse.}
	Neural Cleanse~\cite{DP:wang2019neural_cleanse} detects whether a trained model has been planted backdoor, in which case it assumes the training samples will require minimal modifications to be manipulated by the attacker. Here, we apply Neural Cleanse to detect a backdoored ResNet-34 model by our \proposed~ on GTSRB dataset. As shown in the right subfigure of Fig.~\ref{fig:resist_defence}, Neural Cleanse fails to detect the backdoored model, \ie, anomaly index $<$ 2. More results on other datasets can be found in Appendix \ref{appendix_d}.
	
	\noindent\textbf{Resistance to white-box trigger removal.}
	Here, we apply trigger removal methods for different backdoor attacks in a white-box setting (the defender has identified the trigger pattern). For our \proposed, many reflection removal methods~\cite{DLA:liu2019semantic,DLA:liu2020separate,RI:Zhang2018Perceptual} can be applied. In our experiment, we adopt the state-of-the-art reflection removal method ~\cite{RI:Zhang2018Perceptual} to clean the poisoned data.
	For Badnets, we simply replace the value of the trigger by the mean pixel value of their three adjacent patches.
	For CL, we use the non-Local means denoising technique~\cite{RI:Antoni2011Non}.
	For SIG, we add $-v(i, j)$ (defined in Eqn.~\eqref{eq:sin_func} in Appendix \ref{appendix_d}) to backdoored images to remove the trigger. The attack success rates before and after trigger removal are reported in Table~\ref{tab:trigger_removal}. Existing attacks Badnets, CL, and SIG rely on fixed backdoor patterns, thus can be easily removed by white-box trigger removal methods, \ie, success rate drops to $<20\%$. Conversely, our \proposed~ uses reflection images randomly selected from the wild, thus can still maintain a high success rate of 85\% after reflection removal.
	Overall, we believe backdoor attack is still a challenging task to successfully attack a model while evade white-box trigger removal. Detailed experimental settings and more results on other defenses including input denoising and mixup data augmentation can be found in Appendix \ref{appendix_d}.

	\begin{table}[t]
		\begin{center}
			\caption{The attack success rate (\%) before/after white-box trigger removal on GTSRB dataset.}
			\label{tab:trigger_removal}
			\setlength{\tabcolsep}{3mm}
			{\small
				\begin{tabular}{lllll}
					\bottomrule[1.2pt]
					\specialrule{0em}{1pt}{1pt}
					& Badnets~\cite{DP:gu2017badnets} & CL~\cite{DP:turner2019cleanlabel} & SIG~\cite{DP:barni2019new} & \proposed \\
					\hline
					\specialrule{0em}{1pt}{1pt}
					Before         		& 24.12  & 78.03  & 73.26  & \textbf{91.67} \\
					After & 15.38 \mydarkred{$\blacktriangledown$ 8.74} & 18.18 \mydarkred{$\blacktriangledown$ 59.85} & 17.29 \mydarkred{$\blacktriangledown$ 55.97}  & \textbf{85.01} \mydarkred{$\blacktriangledown$ \textbf{6.65}} \\
					\bottomrule[1.2pt]
			\end{tabular}  }
		\end{center}
	\end{table}

	
	\section{Conclusion} \label{sec:conclusion}
	In this paper, we have explored the natural phenomenon of reflection, for use in backdoor attack on DNNs. Based on mathematical modeling of physical reflection models, we proposed the \emph{reflection backdoor} (\proposed~) approach.  \proposed~ plants a backdoor into a victim model by generating and injecting reflections into a small set of training data. Empirical results across 3 computer vision tasks and 5 datasets demonstrate the effectiveness of \proposed. It can attack state-of-the-art DNNs with high success rate and small degradation in clean accuracy. Reflection backdoors can be generated efficiently, and are resistant to state-of-the-art defense methods.
	It is an open question as to whether new types of training strategies can be developed that are robust to this kind of natural backdoors.
	
	%
	%
	\bibliographystyle{splncs04}
	\bibliography{egbib}
	
	\newpage
	\appendix
	
	\section{Real-world reflections in natural images}
\begin{figure}[htbp]
		\begin{center}
			\includegraphics[width=\linewidth]{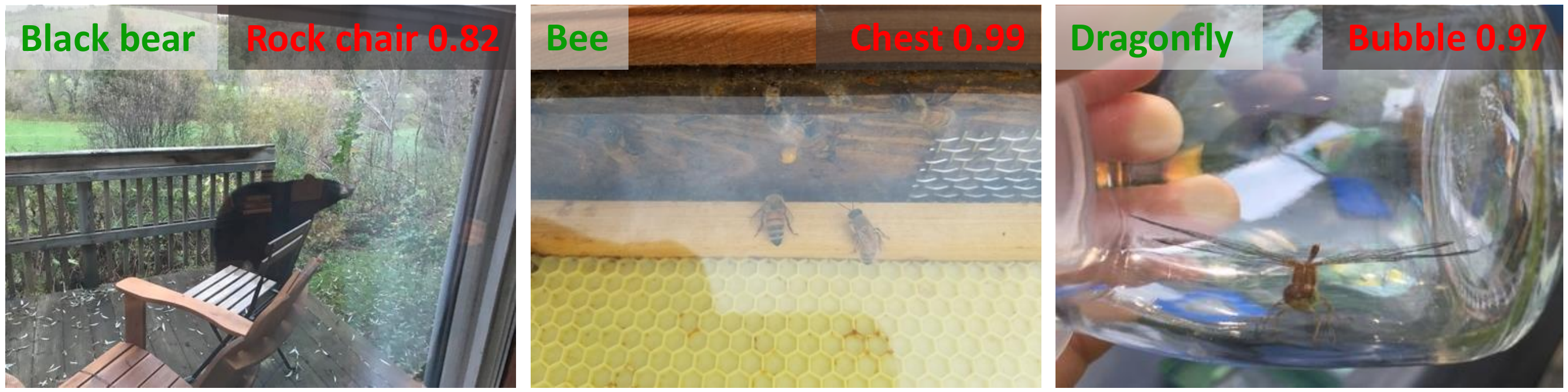}
		\end{center}
		\caption{Real-world reflections (from \textit{ImageNet-a}~\cite{AT:hendrycks2019nae}) influence the performance of a DNN classifier. Labels in green and red colors are ground-truth and predicted labels, respectively.
		}
		\label{fig:2-motivation}
	\end{figure}

	Reflections exist in natural images can also deteriorate classification performance. Fig.~\ref{fig:2-motivation} shows three such examples in the \textit{ImageNet-a}~\cite{AT:hendrycks2019nae} dataset, where all the three images were misclassified by a DNN classifier. For instance, the black bear in the first image was misclassified to be rock chair with 82\% confidence.
	
	\section{More implementation details}
	The statistics of the datasets and DNN models used in our experiments are summarized in Table~\ref{tab:exp_setup}.
	
	\begin{table}[htbp]
		\begin{center}
			\caption{Statistics of image datasets and DNN models used in our experiments.}
			\label{tab:exp_setup}
			\setlength{\tabcolsep}{2.0mm}
			{
				\begin{tabular}{cccccc}
					\bottomrule[1.2pt]
					\specialrule{0em}{1pt}{1pt}
					Task &    Dataset & \# Labels & \tabincell{c}{\# Input\\ Size} & \tabincell{c}{\# Training\\Images}  & \tabincell{c}{DNN\\model}\\
					\hline
					\specialrule{0em}{1pt}{1pt}
					\multicolumn{ 1}{c}{Traffic} &      GTSRB &         13 &  224$\times$224 &       4772 &  ResNet-34 \\
					\multicolumn{ 1}{c}{Sign} & BelgiumTSC &         11 &  224$\times$224 &       3556 &  ResNet-34 \\
					\multicolumn{ 1}{c}{Recognition} &      CTSRD &         22 &  224$\times$224 &       2028 &  ResNet-34 \\
					\hline
					\specialrule{0em}{1pt}{1pt}
					
					\tabincell{c}{Face\\Recognition} & \tabincell{c}{PubFig}  &         60 &  300$\times$300 &       5181 &  ResNet-34 \\
					\hline
					\specialrule{0em}{1pt}{1pt}
					\tabincell{c}{Object\\Classification} & \tabincell{c}{ImageNet\\subset}  &         12 &  300$\times$300 &      12406 & \tabincell{c}{ResNet-34\\DenseNet-121} \\
					\bottomrule[1.2pt]
			\end{tabular}  }
		\end{center}
	\end{table}  
	
	\noindent\textbf{Detailed implementation of baselines.}\label{appendix_a}
	There are two baselines for our experiments.
	For clean-label attack (CL)~\etal~\cite{DP:turner2019cleanlabel}, we use the same settings as reported in their paper. Specifically, we use Projected Gradient Descent (PGD) adversarial perturbation bounded to $L_{\infty}$ maximum perturbation $\epsilon$=16.
	For SIG~\cite{DP:barni2019new}, Backdoored image are generated with horizontal sinusoidal signal defined by 
	\begin{equation} \label{eq:sin_func}
	v(i, j) = \Delta sin(2\pi jf / m), 1 \leq j \leq m, 1 \leq i \leq l,
	\end{equation}
	where $f$ is a certain frequency, we follow~\cite{DP:barni2019new} and set $\Delta = 20$ and $f = 6$. 
	
	\section{Results on more target classes}\label{appendix_b}
	We run more experiments with different target classes (\eg class indexes 1, 2, 3, 4) on GTSRB dataset. The test accuracy and attack success rate are reported in Table~\ref{tab:more_classes_cmp}. While there are some variations, the overall results of our \proposed~attack are consistent over different target classes. 
	
	\begin{table}[htbp]
		\begin{center}
			\caption{Attack success rate and test accuracy (on clean test samples) of our \proposed~attack on different target classes of the GTSRB dataset.}
			\label{tab:more_classes_cmp}
			\setlength{\tabcolsep}{2mm}
			{
				\begin{tabular}{ccc}
					\bottomrule[1.2pt]
					\specialrule{0em}{1pt}{1pt}
					Class ID & Test accuracy & Attack success rate \\
					\hline
					\specialrule{0em}{1pt}{1pt}
					0 & 86.30\% & 91.67\% \\
					1 & 81.75\% & 87.98\% \\ 
					2 &	85.48\% & 89.74\% \\
					3 & 85.75\%	& 90.83\% \\
					4 &	81.29\% & 91.81\% \\
					\bottomrule[1.2pt]
			\end{tabular}  }
		\end{center}
	\end{table} 
	
	\section{More quantitative results for stealthiness comparison}\label{appendix_c}
	
\begin{figure*}[!htb]
		\centering
		\includegraphics[width=0.85\linewidth]{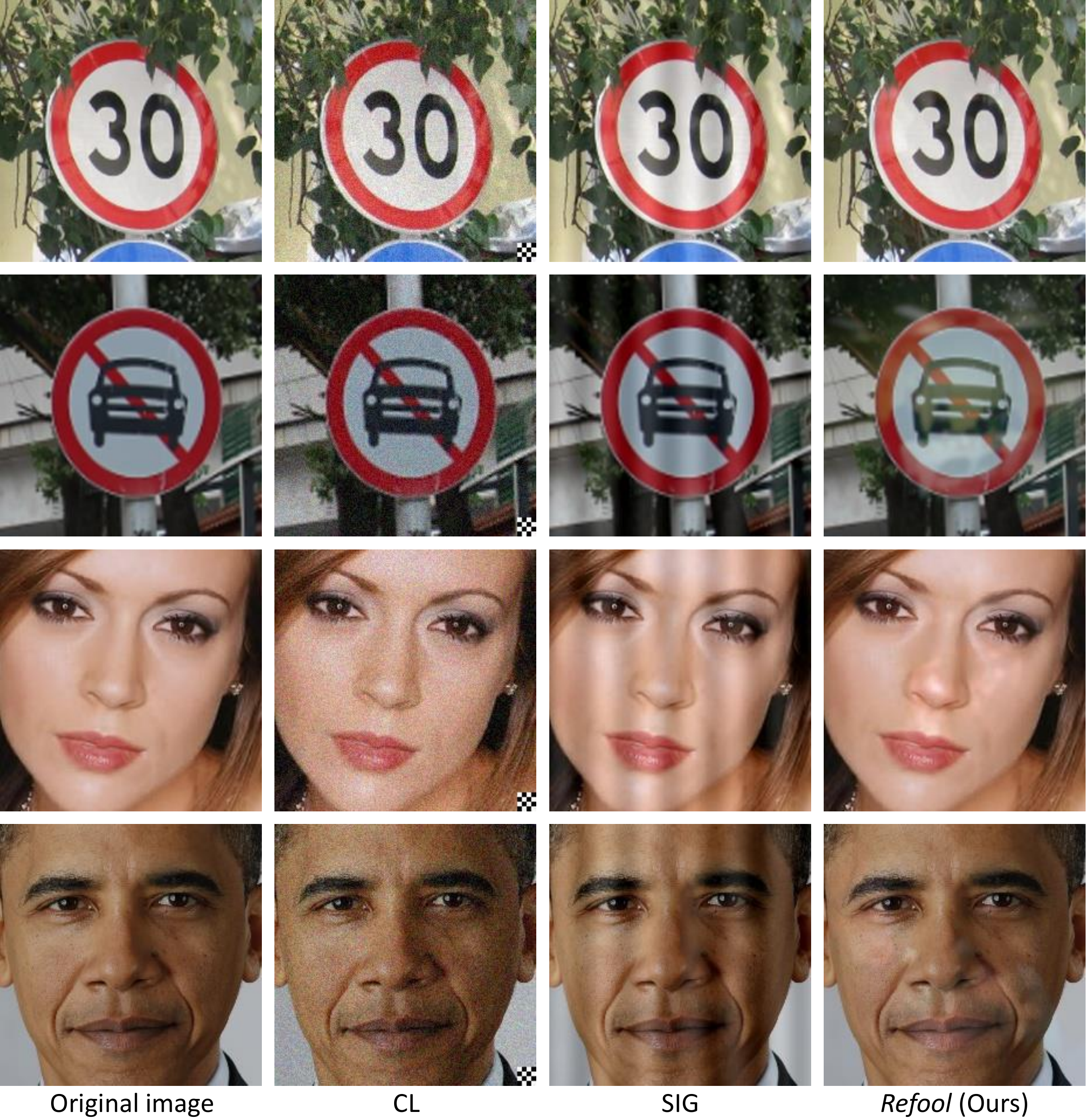}
		\caption{More visual inspections for the stealthiness of CL~\cite{DP:turner2019cleanlabel}, SIG~\cite{DP:barni2019new} and our \proposed.}
		\label{fig:stealthy_cmp_supp}
	\end{figure*}

	By randomly selecting 500 images from CTSRD, we conduct a quantitative comparison of the stealthiness between our \proposed~ and the baselines CL~\cite{DP:turner2019cleanlabel} and SIG~\cite{DP:barni2019new}.
	The average L2, L1 distances and Mean Square Error (MSE) between the original images and their backdoored versions are reported in Table~\ref{tab:stealthy_cmp}. The distortions of our \proposed~ are much lower than either CL or SIG, indicating higher stealthiness. This is further verified by more visual inspections on some randomly selected examples in Fig.~\ref{fig:stealthy_cmp_supp}.
	
	\begin{table}[htbp]
		\begin{center}
			\caption{The average distortions (measured by L2, L1 and MSE distances) made by different backdoor attacks on 500 randomly selected clean training images.}
			\label{tab:stealthy_cmp}
			\setlength{\tabcolsep}{2mm}
			{
				\begin{tabular}{cccc}
					\bottomrule[1.2pt]
					\specialrule{0em}{1pt}{1pt}
					& CL~\cite{DP:turner2019cleanlabel} & SIG~\cite{DP:barni2019new} & \proposed \\
					\hline
					\specialrule{0em}{1pt}{1pt}
					L2 norm &  145.15 & 147.13 & \textbf{113.67} \\
					L1 norm &  119.65 & 125.50 & \textbf{72.06 }\\
					MSE &  273.73 & 201.55 & \textbf{75.30} \\
					\bottomrule[1.2pt]
			\end{tabular}  }
		\end{center}
	\end{table} 
	
	\section{More results against state-of-the-art backdoor defenses}\label{appendix_d}
	\begin{figure*}[!htb]
		\centering
		
		\includegraphics[width=0.75\linewidth]{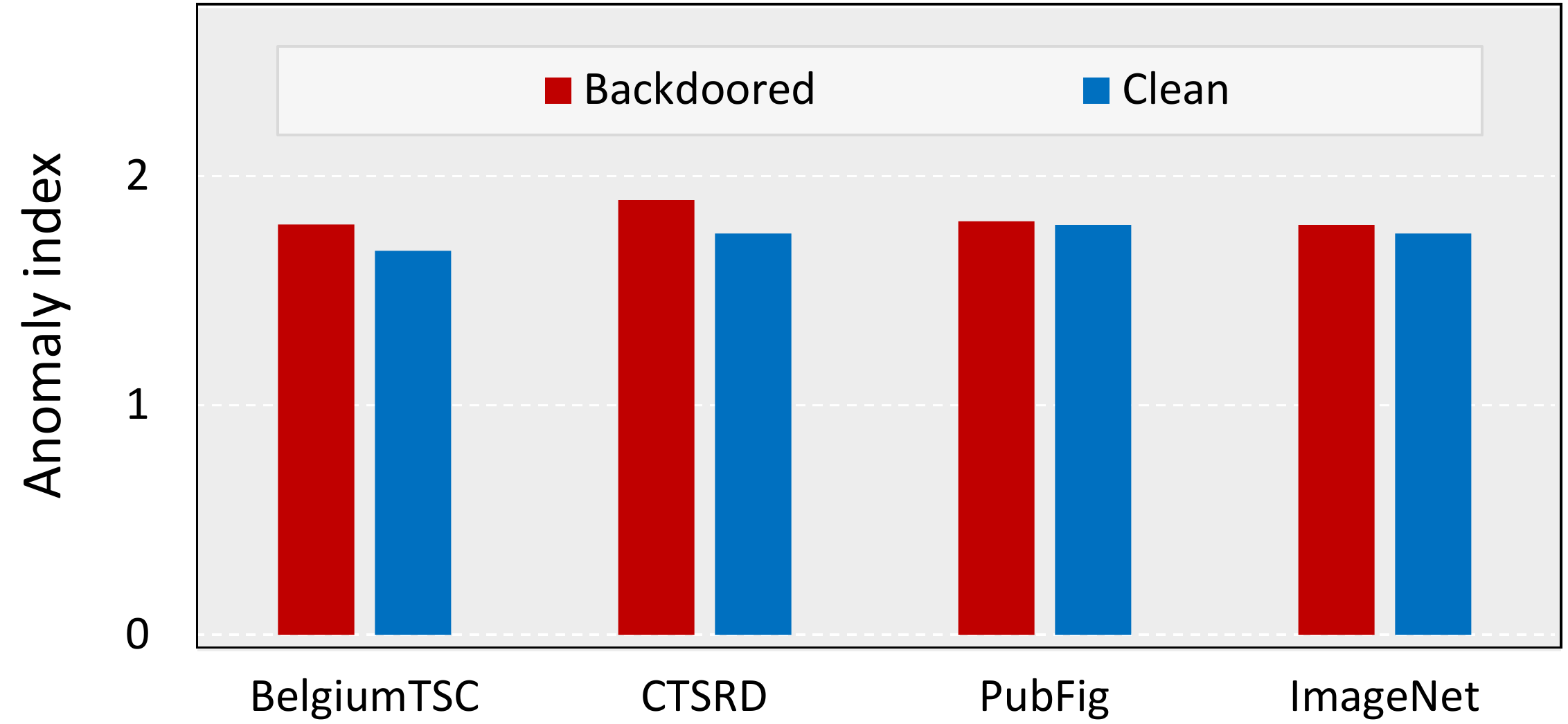}
		
		\caption{More results of Neural Cleanse on five datasets.}
		\label{fig:more_neueal_cleanse}
	\end{figure*}
	
	\noindent\textbf{White-box trigger removal.}
	For Fine-Pruning~\cite{DP:liu2018fine_pruning}, we replicate the Fine-pruning via PyTorch~\cite{ML:PyTorch} and prune the last convolutional layer (\ie, \texttt{layer4.2.conv2}) of the DNNs. In terms of white-box trigger removal, for our \proposed, we adopt a state-of-the-art reflection removal method ~\cite{RI:Zhang2018Perceptual}. For Badnets~\cite{DP:gu2017badnets}, we simply replace the value of the trigger by the mean pixel value of their three adjacent patches. For CL~\etal~\cite{DP:turner2019cleanlabel}, we use the non-Local means denoising technique~\cite{RI:Antoni2011Non}. For SIG~\cite{DP:barni2019new}, we add the $-v(i, j)$ defined in Eqn.~\eqref{eq:sin_func} on backdoored image back to the backdoor image to remove the trigger pattern. We apply trigger removal on the poisoned training data, then retrain the model under the same condition for all the other four datasets: BelgiumTSC, CTSRD, PubFig, and ImageNet.
	As shown in Table~\ref{tab:trigger_removal_supp}, our \proposed~ maintains a much higher success rate after trigger removal than either CL or SIG across all datasets. We notice that \proposed~ also exhibits an obvious success rate drop on ImageNet datasets. We suspect this is caused by the large amount of natural noise exists in ImageNet images. These natural noise tends to affect the effectiveness of all backdoor patterns, and also increase the possibility for them to be removed. We believe that, for our attack, this can be addressed by simply increasing the intensity of the reflection. A more adaptive reflection backdoor to this situation is an interesting future work.
	
	\noindent\textbf{Neural Cleanse detection.}
	Fig.~\ref{fig:more_neueal_cleanse} illustrates more results of \proposed~ backdoored models against Neural Cleanse detection on datasets BelgiumTSC, CTSRD, PubFig and ImageNet. None of the four backdoored models by our \proposed~ can be detected by Neural Cleanse. Note that only an anomaly index $>2$ indicates a successful detection.

	\begin{table}[t]
		\begin{center}
			\setlength{\tabcolsep}{1.5mm}
			\caption{The attack success rate (\%) of different backdoor attacks before or after white-box trigger removal.}
			\label{tab:trigger_removal_supp}
			{
				\begin{tabular}{l|cc|cc|cc|cc}
					\bottomrule[1.2pt]
					\specialrule{0em}{1pt}{1pt}
					Dataset & \multicolumn{2}{c|}{Badnets~\cite{DP:gu2017badnets}} & \multicolumn{2}{c|}{CL~\cite{DP:turner2019cleanlabel}} & \multicolumn{2}{c|}{SIG~\cite{DP:barni2019new}} & \multicolumn{2}{c}{Ours} \\
					& before & after & before & after & before & after & before & after \\
					\hline
					\specialrule{0em}{1pt}{1pt}
					BelgiumTSC    		& 11.40 & 0.75  & 46.25 & 8.33  & 51.86 & 0.88 & 85.70 & 77.78 \\
					CTSRD         		& 25.24 & 7.23  & 63.63 & 11.52 & 57.39 & 6.10 & 91.70 & 83.09 \\
					PubFig        		& 42.86 & 13.33 & 78.67 & 31.74 & 69.01 & 8.34 & 81.30 & 68.42 \\
					ImageNet+ResNet   	& 15.77 & 8.98  & 55.38 & 17.69 & 63.84 & 8.45 & 82.11 & 36.93 \\
					ImageNet+DenseNet 	& 20.14 & 7.32  & 67.43 & 12.93 & 68.00 & 7.37 & 75.16 & 28.07 \\
					\bottomrule[1.2pt]
			\end{tabular}  }
		\end{center}
	\end{table}
	
	\noindent\textbf{Input denoising or data augmentation based defenses.}
	We further evaluated the resistance of our \proposed~ attack to input denoising methods on CTSRD dataset. Specifically, we consider denoising techniques from Guo~\etal~\cite{DP:guo2017input_transformations}: image quilting, Total Variation denoising (TV denoise), JPEG compression, and Pixel quantization. We also include the data augmentation based mixup defense in ~\cite{DP:zhang2017mixup}. These denoising or augmentation defenses are mostly proposed for adversarial attacks, but can be directly applied to backdoor attacks. We apply the denoising methods on all test samples (both backdoored and non-backdoored), and report the model's performance on denoised samples. For mixup, we retrain the network on the backdoored training set with its default setting. As shown in Table~\ref{tab:more_defense_exp}, these denoising or augmentation methods indeed can decrease the attack success rate for ~4\%. However, they are less effective than defenses like fine-tuning or trigger (\eg~ reflection) removal. And image quilting seems greatly decrease the model's performance on clean samples, \ie, test accuracy drops from 86.30\% to 11.35\%.
	
	\begin{table}[t]
		\begin{center}
			\setlength{\tabcolsep}{1.5mm}
			\caption{The resistance of our  \proposed~ attack to input denoising or data augmentation defenses on CTSRD dataset}
			\label{tab:more_defense_exp}
			{
				\begin{tabular}{lcc}
					\bottomrule[1.2pt]
					\specialrule{0em}{1pt}{1pt}
					Methods & Test accuracy (\%) & Attack success accuracy (\%) \\
					\hline
					\specialrule{0em}{1pt}{1pt}
					\textit{Refool} (proposed)  & 86.30 & 91.67 \\
					\hline
					\specialrule{0em}{1pt}{1pt}
					Quilting           & 11.35 & 89.09 \\
					TV denoise         & 85.43 & 89.84 \\
					JPEG compression   & 86.57 & 90.98 \\
					Pixel quantization & 86.30 & 91.01 \\
					Mixup              & 87.79 & 87.08 \\
					Reflection removal & 86.41 & 85.01 \\
					
					\bottomrule[1.2pt]
			\end{tabular}  }
		\end{center}
	\end{table}
	
	
\end{document}